\pdfoutput=1

\documentclass[11pt]{article}

\usepackage[]{EMNLP2022}

\usepackage{times}
\usepackage{latexsym}

\usepackage[T1]{fontenc}

\usepackage[utf8]{inputenc}

\usepackage{microtype}

\usepackage{inconsolata}

\usepackage{multirow}
\usepackage{makecell}
\usepackage{booktabs}

\usepackage{graphicx}
\usepackage{float}  
\usepackage{subfigure}  
\usepackage{amsmath}
\usepackage{caption}
\usepackage{float}  
\usepackage{algorithm, algorithmic}
\usepackage{xcolor}
\usepackage[normalem]{ulem}

\title{Learning to Adapt to Low-Resource Paraphrase Generation}
\author{
  Zhigen Li$^{1}$, Yanmeng Wang$^{1}$, Rizhao Fan$^{2}$, Ye Wang$^{1}$\\
  {\bf Jianfeng Li$^{1}$} {\bf and Shaojun Wang$^{1}$} \\
  $^{1}$Ping An Technology, $^{2}$University of Bologna\\
  \texttt{$\{$lizhigen974,wangyanmeng219,wangye430$\}$@pingan.com.cn} \\
  \texttt{rizhao.fan@unibo.it, $\{$lijianfeng777,wangshaojun851$\}$@pingan.com.cn} \\
}

\usepackage{fancyhdr}
\usepackage{lipsum} 

\fancypagestyle{firstpage}{
    \fancyfoot[C]{\footnotesize \textit{Proceedings of the 2022 Conference on Empirical Methods in Natural Language Processing}, pages 1014 - 1022 \\
    December 7-11, 2022 \textcopyright 2022 Association for Computational Linguistics}
}

\begin{document}

\thispagestyle{firstpage}

\maketitle
\begin{abstract}
Paraphrase generation is a longstanding NLP task and achieves great success with the aid of large corpora. 
However, transferring a paraphrasing model to another domain encounters the problem of domain shifting especially when the data is sparse.
At the same time, widely using large pre-trained language models (PLMs) faces the overfitting problem when training on scarce labeled data. 
To mitigate these two issues, we propose, LAPA, an effective adapter for PLMs optimized by meta-learning. 
LAPA has three-stage training on three types of related resources to solve this problem: 
1. pre-training PLMs on unsupervised corpora, 
2. inserting an adapter layer and meta-training on source domain labeled data, and 
3. fine-tuning adapters on a small amount of target domain labeled data. 
This method enables paraphrase generation models to learn basic language knowledge first, then learn the paraphrasing task itself later, and finally adapt to the target task. 
Our experimental results demonstrate that LAPA achieves state-of-the-art in supervised, unsupervised, and low-resource settings on three benchmark datasets. 
With only 2\% of trainable parameters and 1\% labeled data of the target task, our approach can achieve a competitive performance with previous work.
\end{abstract}

\section{Introduction}
Paraphrase generation can comprehend a sentence and generate another with the same semantics but with variations in lexicon or syntax, which has various applications on downstream tasks including query rewriting~\citep{dong2017learning}, data augmentation~\citep{iyyer2018adversarial} and language model pre-training~\citep{lewis2020pre}.  
Conventional approaches~\citep{prakash2016neural,chowdhury2022novelty} model the paraphrase generation as a supervised encoding-decoding problem, inspired by machine translation systems. 
However, the success of these methods often relies on a large number of parallel paraphrases, whose collection is time-consuming and requires a lot of domain knowledge. 
Therefore, in real scenarios with a small amount of parallel data, the model suffers from performance drops facing domain gaps.
This phenomenon, known as domain shift problem~\citep{pan2009survey}, comes from the representation gap between training and testing domains with different writing styles or forms. 

To tackle this problem, unsupervised methods such as editing-based approaches~\citep{bowman2015generating,miao2019cgmh} or reinforcement learning~\citep{li2017paraphrase,siddique2020unsupervised}, and weakly-supervised methods such as retrieval-enhanced~\citep{ding2021learning, yin2022learning} or prompt-based~\citep{wang2022promda} do not introduce or only introduce a small number of supervised signals, which limits their performance such that underperforms supervised methods.
In fact, large-scale unlabeled corpus data (UCD) and labeled source domain data (LSDD), as well as a few labeled target domain data (LTDD), can be easily achieved. 
Therefore, we propose a new three-stage learning paradigm: pre-training, meta-learning, and fine-tuning, aiming to leverage the pre-trained knowledge on UCD, source domain knowledge on LSDD, and adapt to target domain on LSDD to improve the performance of low-resource paraphrase generation. 
\begin{figure*}[t]
    \vskip -2em 
    \centering
    \includegraphics[width=1.0\linewidth]{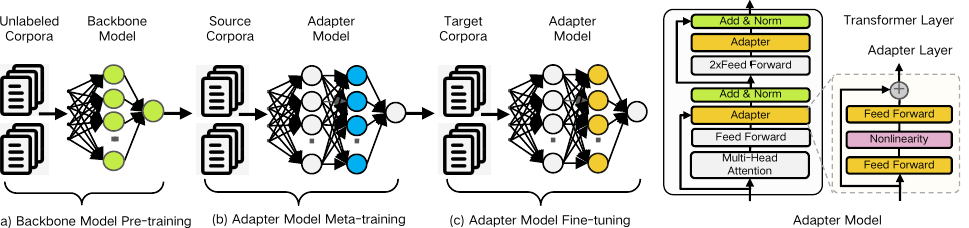}
    \vskip -0.5em
    \caption{Three training stages of the proposed learning paradigm. Gray represents untrainable parameters, and other bright colors represent parameters that have been trained in different stages.}
    \label{fig: Workflow of LAPA}
    \vskip -1.5em
\end{figure*}
In order to successfully implement this learning paradigm, we propose a simple yet effective model which combined pre-trained language model (PLM) and MAML~\citep{finn2017model}, named \textbf{L}earning to \textbf{A}dapt to low-resource \textbf{PA}raphrase generation (\textbf{LAPA}). 
Specifically, before meta-learning, we insert an adapter layer into each transformer layer of PLM. An adapter layer is composed of a few parameters of feed-forward layer and residual connection. 
During meta-training and fine-tuning, only the adapter layer and normalization layer are trainable. 
Parameter freezing and residual connection can retain the prior knowledge of PLM to avoid negative transfer effects. 
Smaller-scale parameter updating can prevent MAML from gradient explosion or diminishing problems when the number of MAML inner loop iterations and model depth increase~\citep{antoniou2018train} or training data is extremely scarce. 

Overall, we hold the idea that paraphrasing is a fundamental ability of human beings. The paraphrase model should not rely on domain and seen data. Therefore, we are committed to characterizing the basic ability of the paraphrase model, obtaining gains from each domain, and applying it to a specific domain. Our contributions are summarized as follows:
\begin{itemize}
	\item We define a novel three stages learning paradigm for low-resource paraphrase generation in data scarcity scenarios.
	\item We propose that LAPA implement this learning paradigm, which transferred the PLM knowledge and source domain knowledge to complete the low-resource learning in the target domain quickly and with high quality.
	\item The supervised, unsupervised and weakly supervised experimental results of LAPA on three benchmark datasets achieve state-of-the-art (SOTA). LAPA with only 2\% of trainable parameters and 1\% target task labeled data can achieve a competitive performance with previous works.
\end{itemize}
\section{Related Work}
While the paraphrase generation performance is greatly improved with various supervised techniques~\citep{zhao2008combining,prakash2016neural,egonmwan2019transformer,cao2020divgan,hosking2021factorising,chowdhury2022novelty}, there are few studies regarding the low-resource setting. ~\citet{west2020reflective} and ~\citet{meng2021conrpg} proposed novel unsupervised paraphrasing strategies by data augmentation based on reflective decoding or diverse decoding. 
~\citet{ding2021learning} and ~\citet{yin2022learning} achieved improvements on various low-resource datasets with retrieved data and meta reinforcement learning. 
However, these studies only use a single large corpus for training the full PLM, which suffers from domain-shifting problems~\citep{wang2019characterizing}. 
Besides, under the extreme low-resource setting, directly fine-tuning the full PLM will cause an over-fitting problem~\citep{antoniou2018train}. 

Meta-learning helps improve low-resource performance in various recent studies, such as image classification~\citep{soh2020meta}, vehicle tracking~\citep{song2020meta} and natural language processing~\citep{park2020unsupervised, chen2021meta, hong2022lea}.
\citet{finn2017model} proposed a meta learner named MAML, which uses other example tasks to learn how to effectively initialize a basic learner, which can be quickly generalized to new tasks. 
Adapter modules have been mainly used for parameter-efficient and quick fine-tuning of a basic PLMs to new tasks~\citep{houlsby2019parameter, bapna2019simple, pfeiffer2020mad, pfeiffer2020adapterfusion, he2021towards}.
Our paper proposes to incorporate meta-learning approaches to realize multi-domain migration and task adapter to realize parameter effective transfer learning (i.e., limited trainable parameters) to mitigate the above problems of paraphrase generation.
\section{The Approach}
\subsection{Learning Paradigm} 
As shown in Figure \ref{fig: Workflow of LAPA}, the workflow of our learning paradigm including three stages: 1. Backbone model pre-training on large unlabeled corpora 2. Adapter model meta-training on large source corpora using the meta-learning and 3. Adapter model fine-tuning on target corpora and evaluate model performance.
The prior knowledge $ K_{pri} $ comes from first two stages: pre-training and meta-learning. 
We denote our backbone model by $f(\theta)$ with parameters $ \theta $. 
The first stage is pre-training on unlabeled corpora $\mathcal D_{pre}$, and we get  $f(\theta_{pre})$. 
The second stage is meta-training on adapter model $f[\theta_{pre},\Phi]$ with additional parameters $ \Phi $ and frozen $ \theta_{pre} $ on related source corpora $ \mathcal D_{src} $, and we got $f[\theta_{pre},\Phi_{src}]$.
Finally, we initialize the adapter model with $[\theta_{pre},\Phi_{src}]$ and fine-tune $\Phi_{src}$ on the target corpus $\mathcal D_{tgt} $ to obtain a target model $f[\theta_{pre}, \Phi_{tgt}]$ which are model parameters after target adapter, i.e., the posterior knowledge $K_{por} $.
\subsection{Backbone Model}
Because PLM is equipped with prior knowledge $ K_{pri} $ and exhibits strong capabilities in a range of different generative tasks, we choose the pre-trained BART~\citep{lewis2019bart} as the backbone model for paraphrase generation. Specifically, given a labeled paraphrase pair $i=(\mathbf{x}, \mathbf{\hat{y}})$, where $ \mathbf{x} = [x_{1} , \ldots, x_{N}] $, $ \mathbf{\hat{y}}=[\hat{y}_{1} , \ldots, \hat{y}_{M} ]$, and inputting $ \mathbf{x} $, the model has produced a predicted segment sequence $ \mathbf{y}_{<t} = [y
_{1} , \ldots, y_{t-1}]$ before time $t$, then the probability that the token generated at time $t$ is $y_{t}$ is $ p(y_{t}|\mathbf{y}_{<t},\mathbf{x},\theta)$.
The model is optimized by minimizing the negative log-likelihood:
$
\label{math loss}
\mathcal L_{i}(f(\theta)) = -\sum_{t=1}^{M} log \, p(\hat{y}_{t}|\mathbf{y}_{<t},\mathbf{x},\theta)
$.
\subsection{Adapter Model}

The adapter model is obtained by inserting the adapter layer into each transformer layer of the backbone model.
An adapter layer is a bottle-necked feed-forward network consisting of a down-project layer, a nonlinearity function and an up-project layer. 
In addition, a skip connection layer from input to output prevents the noised initialization from interference with the training initially. 
For the adapter in layer $l$, the function can be formulated as:
$
\label{math adapter}
Adapter(\mathbf{z}_{l}) = \mathbf{W}^{l}_{u}ReLU(\mathbf{W}^{l}_{d}\mathbf{z}_{l}) + \mathbf{z}_{l} 
$
where $z_{l}$ represents the inputs of the adapter in layer $l$. 
Besides, the normalization layers are trainable and initialized from the previous training stage.
\subsection{Meta-Learning}
\begin{algorithm}
	\caption{Adapter Model Training with  Model Agnostic Meta-Learning}
	\label{algorithm mtl}
	\begin{algorithmic}[1]
		\REQUIRE $p(\mathcal{T})$: distribution over tasks; stage (b) over  $\mathcal{D}_{src}$, and stage (c) over $\mathcal{D}_{tgt}$
		\REQUIRE $f[\theta,\Phi]$: adapter model
		\REQUIRE $\theta_{pre}$: pre-trained backbone model parameters
		\REQUIRE $\Phi_{init}$: initialization parameters of adapters; stage (b) is the zero, and stage (c) is $\Phi_{src}$ learned from $\mathcal{D}_{src}$
		\REQUIRE $\alpha,\beta$: step size hyperparameters
		\STATE Initialize $[\theta,\Phi] \gets [\theta_{pre}, \Phi_{init}]$
		\STATE Fix $\theta$ in the training procedure
		\WHILE{not done}
		    \STATE Sample batch of tasks $\mathcal{T}_{i} \sim p(\mathcal{T})$
		    \FOR{all $\mathcal{T}_{i}$}
    \STATE Evaluate gradient $\nabla_{\Phi}\mathcal{L}_{i}(f[\theta,\Phi])$ with respect to $K$ examples
    \STATE Compute adapted parameters with gradient descent: 
    $[\theta, \hat{\Phi}] = [\theta, \Phi] - \alpha \nabla_{\Phi}\mathcal{L}_{i}(f[\theta,\Phi])$
            \ENDFOR
            \STATE Update $ [\theta, \Phi] \gets [\theta, \Phi] - \beta \nabla_{\Phi} \sum_{\mathcal{T}_{i} \sim p(\mathcal{T})} \mathcal{L}_{i}(f[\theta,\hat{\Phi}]) $
		\ENDWHILE
	\end{algorithmic}  
\end{algorithm}
The second stage is adapter model meta traning based on MAML~\citep{finn2017model}.
The learning process is shown in Algorithm \ref{algorithm mtl}. 
First, we freeze the backbone model parameters $\theta_{pre}$ that have been pre-trained in the pre-training stage, then, add new adapters with parameters $\Phi$ to get adapter model $f[\theta_{pre}, \Phi]$. 
Based on Algorithm \ref{algorithm mtl}, we first complete the meta-learning of the adapter model on the source corpus $\mathcal{D}_{src}$ to help the adapters $\Phi$ find the initialization parameters $\Phi_{src}$  suitable for paraphrase generation to adapt faster target task.
At this time, we obtain the model $f[\theta_{pre}, \phi_{src}]$ with knowledge of the paraphrase generation task. 
In the third stage, we initialize the parameters of adapter model with $[\theta_{pre}, \phi_{src}]$. Then, based on the Algorithm \ref{algorithm mtl}, we fine-tune adapters $\phi$ on target corpus $\mathcal{D}_{tgt}$ to quickly adapt to the target corpus. Finally, we get the target model $f[\theta_{pre}, \phi_{tgt}]$.
\section{Experimental Settings}
\subsection{Datasets}
\label{sec:datasets}
We conducted experiments on Quora\footnote{https://www.kaggle.com/c/quora-question-pairs}, Twitter~\citep{lan2017continuously} and MSCOCO~\citep{lin2014microsoft} benchmark datasets, and followed the same setting in previous works~\citep{lin2014microsoft, liu2019unsupervised,ding2021learning}. 
For meta-learning, we choose a different source task's labeld train-set from the target task to randomly construct meta tasks. Appendix Table~\ref{dataset-statistics-table} describes more details.
\begin{table*}[htb]
\resizebox{\textwidth}{29mm}{
\begin{tabular}{llrrrrrrrr}
\toprule[1.5pt]
\multicolumn{1}{l}{\multirow{2}{*}{}} & 
\multicolumn{1}{l}{\multirow{2}{*}{\textbf{Method}}} & 
\multicolumn{4}{c}{\textbf{Quora}} & 
\multicolumn{4}{c}{\textbf{Twitter}} \\
 \cmidrule(r){3-6} 
 \cmidrule(r){7-10}
\multicolumn{1}{l}{} & 
\multicolumn{1}{c}{} & 
\multicolumn{1}{l}{BLEU-2} & 
\multicolumn{1}{l}{BLEU-4} & 
\multicolumn{1}{l}{ROUGE-1} & 
\multicolumn{1}{l}{ROUGE-2} & 
\multicolumn{1}{l}{BLEU-2} & 
\multicolumn{1}{l}{BLEU-4} & 
\multicolumn{1}{l}{ROUGE-1} & 
\multicolumn{1}{l}{ROUGE-2} \\
\hline
\multirow{8}{*}{\textbf{Supervised}} 
     & Res-LSTM & 38.52 & 24.56 & 59.69 & 32.71 & 32.13 & 25.92 & 41.77  & 27.94  \\
     & Transformer & 42.91 & 30.38 & 61.25  & 34.23  & 40.34 & 32.14 & 44.53  & 29.55  \\
    & RbM         & 43.54 & -     & 64.39  & 38.11  & 44.67 & -     & 41.87  & 24.23  \\
    & RaE         & 40.35 & 25.37 & 62.71  & 31.77  & 44.33 & 34.16 & 47.55  & 31.53  \\
     & FSET        & 51.03 & 33.46 & 66.17  & 39.55  & 46.35 & 34.62 & 49.53  & 32.04  \\
     & ConRPG      & -     & 26.81 & 65.03  & 38.49  & -     & -     & -      & -      \\
     & SGCP-R      & -     & 38.00 & 68.10  & 45.70  & -     & -     & -      & -      \\
     & LAPA (ours) & \textbf{55.61} & \textbf{39.28} & \textbf{70.78} & \textbf{48.27} & \textbf{54.80} & \textbf{42.18} & \textbf{64.14} & \textbf{47.57} \\
     \hline
\multirow{3}{*}{\textbf{Low-Resource}}    & WS-BART & 44.19 & 31.18 & 58.69  & 33.39  & 45.03 & 34.00 & 51.34  & 35.89  \\
     & LTSL       & 49.18 & 36.05 & 64.36  & 39.71  & 49.30 & 37.94 & 56.02  & 40.61  \\
     & MB-RPG     & \textbf{54.88} & \textbf{41.56} & 67.66  & 43.98  & 51.65 & 39.58 & 61.45  & 44.19  \\
     & LAPA (ours) & 54.10 & 37.51 & \textbf{70.35}  & \textbf{47.24}  & \textbf{52.71} & \textbf{40.13} & \textbf{63.12}  & \textbf{46.23}  \\
    \bottomrule[1.5pt]
\end{tabular}
}
\vskip 0.5em
\resizebox{\textwidth}{29mm}{
\begin{tabular}{llrrrrrrrr}
\toprule[1.5pt]
\multicolumn{1}{l}{\multirow{2}{*}{}} & 
\multicolumn{1}{l}{\multirow{2}{*}{\textbf{Method}}} & 
\multicolumn{4}{c}{\textbf{Quora}}       & 
\multicolumn{4}{c}{\textbf{MSCOCO}}      \\
 \cmidrule(r){3-6}
 \cmidrule(r){7-10}
\multicolumn{1}{l}{} & 
\multicolumn{1}{c}{} & 
\multicolumn{1}{l}{iBLEU} & 
\multicolumn{1}{l}{BLEU-4} & 
\multicolumn{1}{l}{ROUGE-1} & 
\multicolumn{1}{l}{ROUGE-2} & 
\multicolumn{1}{l}{iBLEU} & 
\multicolumn{1}{l}{BLEU-4} & 
\multicolumn{1}{l}{ROUGE-1} & 
\multicolumn{1}{l}{ROUGE-2} \\
\hline
\multirow{8}{*}{\textbf{Unsupervised}}        
     & VAE        & 8.16  & 13.96 & 44.55  & 22.64  & 7.48  & 11.09 & 31.78  & 8.66   \\
      & UPSA       & 12.02 & 18.18 & 56.51  & 30.69  & 9.26  & 14.16 & 37.18  & 11.21  \\
     & PUP        & 14.91 & 19.68 & 59.77  & 30.47  & 10.72 & 15.81 & 37.38  & 13.87  \\
     & BackTrans  & 15.51 & 26.91 & 52.56  & 27.85  & 7.53  & 10.80 & 36.12  & 11.03  \\
     & set2seq+RTT            & 14.66 & 22.53 & 59.98  & 34.09  & 11.39 & 17.93 & 40.28  & 14.04  \\
     & ConRPG     & 12.68 & 18.31 & 59.62  & 33.10  & 11.17 & 16.98 & 39.42  & 13.50  \\
     & DBlock     & 20.93 & 26.76 & 65.60  & 42.09  & - & - & -  & -  \\
     & LAPA (ours)          & \textbf{25.53} & \textbf{35.12} & \textbf{68.46}  & \textbf{45.09}  & \textbf{11.96} & \textbf{23.48} & \textbf{52.15}  & \textbf{26.77}  \\
    \hline
\multirow{3}{*}{\textbf{Low-Resource}}    & WS-BART    & 17.04 & 31.18 & 58.69  & 33.39  & 10.91 & 15.90 & 40.65  & 15.62  \\
     & LTSL         & 19.20 & 36.05 & 64.36  & 39.71  & 13.45 & 18.87 & 45.18  & 19.17  \\
     & MB-RPG       & \textbf{33.56} & \textbf{41.56} & 67.66  & 43.98  & \textbf{28.09} & 19.39 & 49.42  & 25.18  \\
     & LAPA (ours)  & 26.62          & 37.51 & \textbf{70.35}  & \textbf{47.24}  & 17.79 & \textbf{23.22} & \textbf{54.93}  & \textbf{28.89} \\ 
\bottomrule[1.5pt]
\end{tabular}}
\caption{
Comparisons of LAPA with baseline methods.
We report average scores across five random seeds.
}
\label{main-result-table}
\vskip -1.0em
\end{table*}
\subsection{Baselines}
\label{sec:baselines}
\noindent 
\textit{Supervised} methods are trained with all parallel sentences of target task. \textit{Unsupervised} baselines do not use any labels of target task. Results of ConRPG and SGCP-R are from ~\citep{meng2021conrpg} and ~\citep{kumar2020syntax}. Results for VAE are copied from ~\citet{meng2021conrpg}. For others, we use previously reported results in ~\citet{ding2021learning} and ~\citet{yin2022learning}.
\textit{Low-resource} methods used a highly small amount training data of target task. 
The baseline models compared include the recent SOTA model LTSL~\citep{ding2021learning} , MB-RPG(contemporaneous with our work)~\citep{yin2022learning} and WS-BART with the full parameter fine-tuning based on BART~\citep{lewis2019bart}. Like our work, they all used BART as PLM.
To compare the performance of our method against the previous works, we use BLEU~\citep{papineni2002bleu}, iBLEU~\citep{sun2012joint} and ROUGE~\citep{hovy2006automated} metrics.
All metrics are computed between the generated and the reference paraphrases in the test set~\citep{kumar2020syntax}.
\section{Experimental Results}
Table \ref{main-result-table} summarizes the performance of different methods on Quora, Twitter and MSCOCO datasets. The best score is shown in bold. 
Overall, our LAPA method achieves SOTA performance on most metrics across multiple datasets and different scene settings, demonstrating the effectiveness of our proposed framework. At the same time, low-resource LAPA also approaches or even exceeds the supervised SOTA method (i.e SGCP-R) (Quora’s BLEU-4 are comparable, and other metrics are exceeded). Compared with supervised methods that need to learn from a large amount of target task labeled parallel paraphrases, our method makes full use of other source domain paraphrases and can achieve comparable results with very low-cost supervision signals in target-domain.

\section{Analyses}
\subsection{Parameter Study}
\begin{figure}[htbp] 
    \vskip -1.0em
	\centering
	\includegraphics[width=\linewidth,scale=1.00]{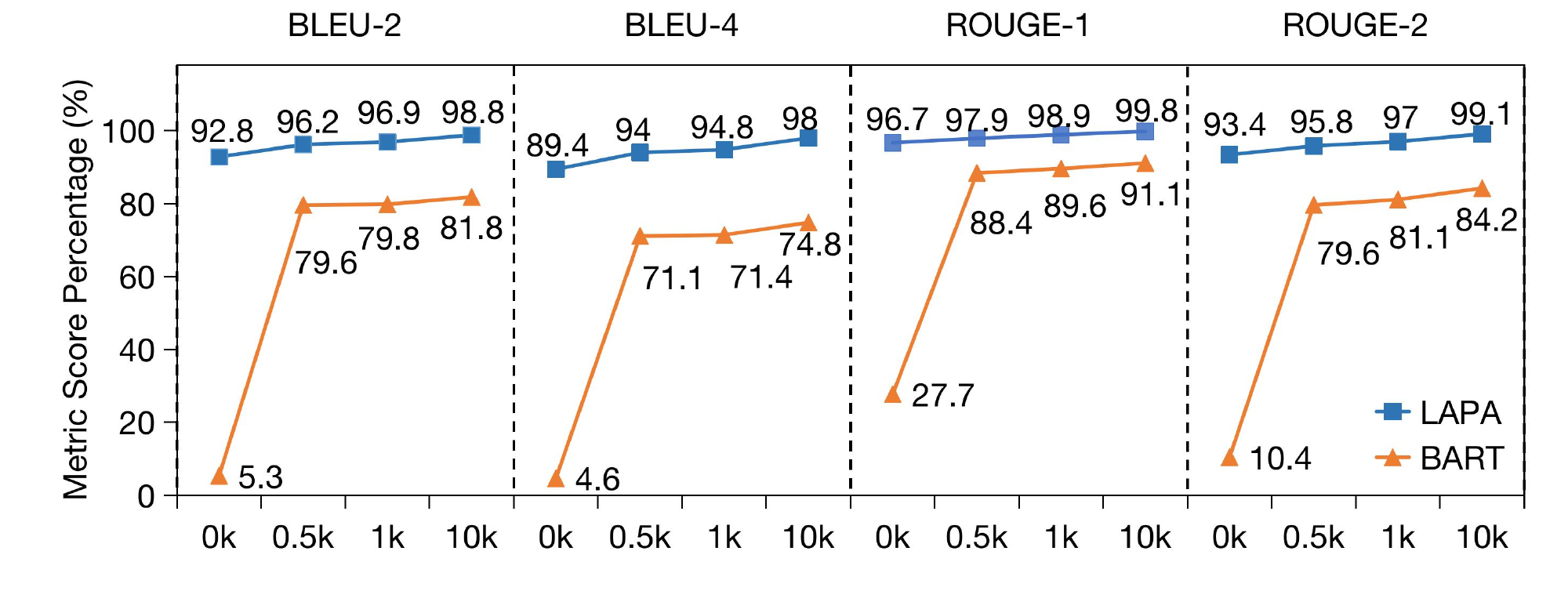}
	\vskip -1em
	\caption{The experimental results of different target data size on Quora for low-resource setting.}
	\label{data_size_fig}
	\vskip -0.5em
\end{figure}
\begin{table*}[htbp]
\vskip -1.5em
\resizebox{\textwidth}{23mm}{
\begin{tabular}{m{2.5cm}m{2cm}m{5cm}m{6cm}m{8cm}}
\toprule[1.5pt]
    {\multirow{2}{*}{\textbf{Example}}} & 
    \textbf{Input} &  
    Can we ever store energy produced in lightning? & 
    How does a pencil and a liquid eyeliner differ? & 
    How come there's no physical evidence for sea dragons existing if they're the largestanimal in the sea. \\ 
\cline{2-5} & 
    \textbf{Reference} & 
     \uwave{Can we store the energy from lightining}? & 
     \uwave{What is the difference between a liquid eyeliner and a pencil} eyeliner? & 
     \uwave{Why is there no evidence of sea dragons existing}? \\ 
    \hline
    {\multirow{2}{*}{\textbf{Unsupervised}}} & 
    \textbf{BART} & 
    CanCan we ever \textcolor{blue}{store energy} & How & How. \\ 
\cline{2-5}
    & \textbf{LAPA} & 
     \uwave{Can we }\textcolor{magenta}{\uwave{store energy from lightning}}? & 
     \uwave{What is the }\textcolor{magenta}{\uwave{difference between liquid eyeliner and a pencil}}? & 
    How come \qquad \uwave{there }\uwave{is }\textcolor{magenta}{\uwave{no}} \textcolor{magenta}{ physical} \textcolor{magenta}{\uwave{evidence of sea dragons}}? \\ 
    \hline
      {\multirow{2}{*}{\textbf{Low-Resource}}} & \textbf{BART} & 
    Is \textcolor{orange}{lightning energy storeable}? & 
    What is the \textcolor{orange}{difference between a liquid eyeliner and a pencil}? & 
    What is the \textcolor{orange}{physical evidence that sea dragons exist}? \\ 
\cline{2-5}
         & \textbf{LAPA} & 
        \uwave{Can we }\textcolor{teal}{\uwave{store energy} produced by \qquad \uwave{lightning}}? & 
        How does \textcolor{teal}{\uwave{a liquid eyeliner differ} from \qquad \uwave{a pencil}}? & 
        How can \textcolor{teal}{\uwave{sea dragons exist} if they're the largest animal in the sea}?. \\ 
        \hline
        {\multirow{2}{*}{\textbf{Supervised}}} & \textbf{BART} & Is it possible to \textcolor{cyan}{store the energy of lightning}? & 
        How do \textcolor{cyan}{liquid eyeliners and pencil eyeliners differ} from each other? & 
        Why is the \textcolor{cyan}{sea dragon the largest animal} in the world? \\
\cline{2-5}
         & \textbf{LAPA} & 
        \uwave{Can we }\textcolor[RGB]{145,213,66}{\uwave{store energy from lightning}}? & 
        \uwave{What is the }\textcolor[RGB]{145,213,66}{\uwave{difference between liquid eyeliner and a pencil}}? & 
        \uwave{Why is there }\textcolor[RGB]{145,213,66}{\uwave{no} physical \qquad \uwave{evidence of sea dragons existing}}? \\ 
\bottomrule[1.5pt]
\end{tabular}}
\caption{
Examples of the generated paraphrases on Quora dataset. We highlight the key phrases in the paraphrases generated and use wavy underline to show the matched parts between LAPA and reference.
}
\label{case-result-table}
\vskip -1.5em
\end{table*}
\noindent We also separately analyze the impact of target task labeled data scale under low-resource setting. 
Figure \ref{data_size_fig} shows the experimental results on the Quora dataset. 
It can be conclused that LAPA has a significant effect compared with BART under the same small data size. LAPA can achieve the effect of 89\% to 93\% of the full amount of data when not using any target task labeled data; when using a very small amount of data such as 0.5k (i.e 0.5\% of the full data), it can be improved to 94\% to 96\%; when the amount of data increases to 10k (i.e 10\% of the full data), the performance is almost the same as the full amount of data 100k.

\begin{figure}[htbp]   
     \vskip -0.5em
	\centering
	\includegraphics[width=\linewidth,scale=1.00]{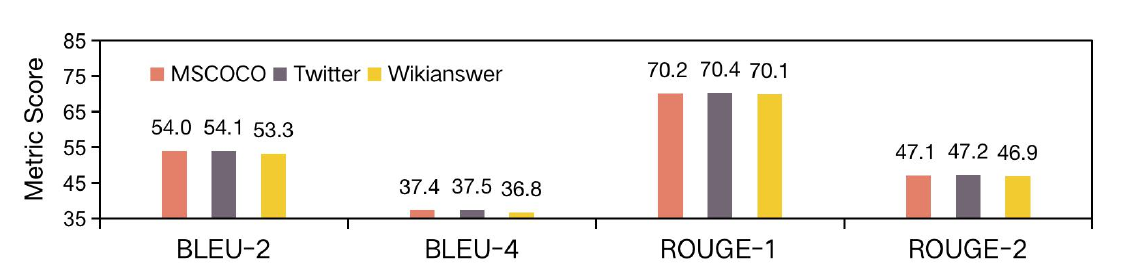}
	 \vskip -1em
	\caption{The experimental results of different source corpus for Quora target task under low-resource setting.}
	\label{FigureCorpus}
	\vskip -0.5em
\end{figure}
It should be pointed out that which dataset is selected as the source data can not have a substantial impact on the migration results, as shown in Figure \ref{FigureCorpus}. 
The results independent of the source dataset prove that LAPA can learn the paraphrasing task itself on any dataset, so it has strong adaptability to the target task.
\subsection{Ablation Study}
\begin{table}[htbp]
\centering
\scalebox{0.65}{
\begin{tabular}{lccccc}
\hline
 Method & T.P. & BLEU-2 & BLEU-4 & Rouge-1 & Rouge-2\\ \hline
BART & 418M &  44.19 & 31.18 & 58.69 & 33.39 \\
BART+SD & 418M & 51.61 & 35.12 & 68.46 & 45.09 \\ \hline
BART+SD+ML & 12M & 54.10 & 37.51 & 70.35 & 47.24 \\ \hline
\end{tabular}}
\caption{
   \label{ablation-table} 
   Ablation results of pre-trained BART, source data (SD) and meta-learning (ML) variants on Quora dataset. 
   T.P. denote trainable parameters.}
   \vskip -1.0 em
\end{table}
We conduct an ablation study with three variants under the low-resource setting of the Quora dataset to investigate the contribution of each component in the proposed method. The experimental results are shown in Table \ref{ablation-table}. We can get: first, using pre-trained BART can get good results; second, by adding the source task dataset for pre-trained BART, the knowledge of the source domain can be effectively learned, thereby improving the performance of the model in the target domain; third, adding our proposed meta-learning framework can again effectively improve the speed and quality of learning the source domain (LAPA only has 2.8\% training parameters compared with BART) and achieve the best performance.
\subsection{Case Study}
Table \ref{case-result-table} lists some paraphrases generated by LAPA and BART with different experimental settings. We can observe that paraphrases produced by LAPA are not only grammatically correct but preserve the semantics of \textit{Input} more completely, and the expression is closer to \textit{Reference} than the other methods.
This benefits from the fact that our LAPA approach can make full use of source domain data and task features, and better preserve the prior knowledge of PLM, so as to adapt to new target tasks quickly and efficiently.
\section{Conclusion}
In this work, we investigate the problem of paraphrase generation under the low-resource setting and propose a simple yet effective approach LAPA. We effectively combine transfer learning and meta-learning by using adapter modules as the bridge. Whether in supervised, unsupervised or low-resource setting, the results that our approach achieves the SOTA results on benchmark datasets. In the future, we plan to explore how to choose a smaller but suitable high-quality source corpus for learning in the 
source domain to improve the effect of transferring to the target domain, because not all source domain data has a positive effect.
Second, we plan to extend this framework to other AI fields to solve low-resource problems in other scenarios and enable more industrial applications.
\clearpage
\section*{Limitations}
The major limitation of present study is the need for source domain annotated data that can adapt to the target domain. Because this is the source of data for the knowledge of the learning task itself, it cannot be avoided. In the real world, we can find it from public free datasets, exchange it commercially with other institutions, or annotate a batch of raw data ourselves as a cold start to solve this problem. Secondly, this study also has insufficient research on related variables. Due to the limitation of time and article length, we have not been able to study. These findings provide the following insights for future research: What is the lower bound of the amount of source domain data that can be well adapted to the target task? Whether we can apply weak supervision, data augmentation and other methods to 
create source domain data? How to select high-quality source domain data to get a better adapter model? 
We leave these questions to future research.
\bibliography{custom}

\begin{thebibliography}{44}
\expandafter\ifx\csname natexlab\endcsname\relax\def\natexlab#1{#1}\fi

\bibitem[{Antoniou et~al.(2019)Antoniou, Edwards, and
  Storkey}]{antoniou2018train}
Antreas Antoniou, Harrison Edwards, and Amos~J. Storkey. 2019.
\newblock \href {https://openreview.net/forum?id=HJGven05Y7} {How to train your
  {MAML}}.
\newblock In \emph{7th International Conference on Learning Representations,
  {ICLR} 2019, New Orleans, LA, USA, May 6-9, 2019}. OpenReview.net.

\bibitem[{Bapna and Firat(2019)}]{bapna2019simple}
Ankur Bapna and Orhan Firat. 2019.
\newblock \href {https://doi.org/10.18653/v1/D19-1165} {Simple, scalable
  adaptation for neural machine translation}.
\newblock In \emph{Proceedings of the 2019 Conference on Empirical Methods in
  Natural Language Processing and the 9th International Joint Conference on
  Natural Language Processing (EMNLP-IJCNLP)}, pages 1538--1548, Hong Kong,
  China. Association for Computational Linguistics.

\bibitem[{Bowman et~al.(2016)Bowman, Vilnis, Vinyals, Dai, Jozefowicz, and
  Bengio}]{bowman2015generating}
Samuel~R. Bowman, Luke Vilnis, Oriol Vinyals, Andrew Dai, Rafal Jozefowicz, and
  Samy Bengio. 2016.
\newblock \href {https://doi.org/10.18653/v1/K16-1002} {Generating sentences
  from a continuous space}.
\newblock In \emph{Proceedings of The 20th {SIGNLL} Conference on Computational
  Natural Language Learning}, pages 10--21, Berlin, Germany. Association for
  Computational Linguistics.

\bibitem[{Cao and Wan(2020)}]{cao2020divgan}
Yue Cao and Xiaojun Wan. 2020.
\newblock \href {https://doi.org/10.18653/v1/2020.findings-emnlp.218}
  {{D}iv{GAN}: Towards diverse paraphrase generation via diversified generative
  adversarial network}.
\newblock In \emph{Findings of the Association for Computational Linguistics:
  EMNLP 2020}, pages 2411--2421, Online. Association for Computational
  Linguistics.

\bibitem[{Chen and Shuai(2021)}]{chen2021meta}
Yi-Syuan Chen and Hong-Han Shuai. 2021.
\newblock \href {https://arxiv.org/abs/2102.09397} {Meta-transfer learning for
  low-resource abstractive summarization}.
\newblock \emph{arXiv preprint arXiv:2102.09397}.

\bibitem[{Chowdhury et~al.(2022)Chowdhury, Zhuang, and
  Wang}]{chowdhury2022novelty}
Jishnu~Ray Chowdhury, Yong Zhuang, and Shuyi Wang. 2022.
\newblock \href {https://arxiv.org/abs/2202.00535} {Novelty controlled
  paraphrase generation with retrieval augmented conditional prompt tuning}.
\newblock \emph{arXiv preprint arXiv:2202.00535}.

\bibitem[{Ding et~al.(2021)Ding, Li, Li, Fan, Guo, Liu, and
  Liu}]{ding2021learning}
Kaize Ding, Dingcheng Li, Alexander~Hanbo Li, Xing Fan, Chenlei Guo, Yang Liu,
  and Huan Liu. 2021.
\newblock \href {https://arxiv.org/abs/2109.12457} {Learning to selectively
  learn for weakly-supervised paraphrase generation}.
\newblock \emph{arXiv preprint arXiv:2109.12457}.

\bibitem[{Dong et~al.(2017)Dong, Mallinson, Reddy, and
  Lapata}]{dong2017learning}
Li~Dong, Jonathan Mallinson, Siva Reddy, and Mirella Lapata. 2017.
\newblock \href {https://doi.org/10.18653/v1/D17-1091} {Learning to paraphrase
  for question answering}.
\newblock In \emph{Proceedings of the 2017 Conference on Empirical Methods in
  Natural Language Processing}, pages 875--886, Copenhagen, Denmark.
  Association for Computational Linguistics.

\bibitem[{Egonmwan and Chali(2019)}]{egonmwan2019transformer}
Elozino Egonmwan and Yllias Chali. 2019.
\newblock \href {https://doi.org/10.18653/v1/D19-5627} {Transformer and seq2seq
  model for paraphrase generation}.
\newblock In \emph{Proceedings of the 3rd Workshop on Neural Generation and
  Translation}, pages 249--255, Hong Kong. Association for Computational
  Linguistics.

\bibitem[{Fader et~al.(2013)Fader, Zettlemoyer, and
  Etzioni}]{fader2013paraphrase}
Anthony Fader, Luke Zettlemoyer, and Oren Etzioni. 2013.
\newblock \href {https://aclanthology.org/P13-1158} {Paraphrase-driven learning
  for open question answering}.
\newblock In \emph{Proceedings of the 51st Annual Meeting of the Association
  for Computational Linguistics (Volume 1: Long Papers)}, pages 1608--1618,
  Sofia, Bulgaria. Association for Computational Linguistics.

\bibitem[{Finn et~al.(2017)Finn, Abbeel, and Levine}]{finn2017model}
Chelsea Finn, Pieter Abbeel, and Sergey Levine. 2017.
\newblock \href {http://proceedings.mlr.press/v70/finn17a.html} {Model-agnostic
  meta-learning for fast adaptation of deep networks}.
\newblock In \emph{Proceedings of the 34th International Conference on Machine
  Learning, {ICML} 2017, Sydney, NSW, Australia, 6-11 August 2017}, volume~70
  of \emph{Proceedings of Machine Learning Research}, pages 1126--1135. {PMLR}.

\bibitem[{He et~al.(2021)He, Zhou, Ma, Berg-Kirkpatrick, and
  Neubig}]{he2021towards}
Junxian He, Chunting Zhou, Xuezhe Ma, Taylor Berg-Kirkpatrick, and Graham
  Neubig. 2021.
\newblock \href {https://arxiv.org/abs/2110.04366} {Towards a unified view of
  parameter-efficient transfer learning}.
\newblock \emph{arXiv preprint arXiv:2110.04366}.

\bibitem[{Hong and Jang(2022)}]{hong2022lea}
SK~Hong and Tae~Young Jang. 2022.
\newblock Lea: Meta knowledge-driven self-attentive document embedding for
  few-shot text classification.
\newblock In \emph{Proceedings of the 2022 Conference of the North American
  Chapter of the Association for Computational Linguistics: Human Language
  Technologies}, pages 99--106.

\bibitem[{Hosking and Lapata(2021)}]{hosking2021factorising}
Tom Hosking and Mirella Lapata. 2021.
\newblock \href {https://doi.org/10.18653/v1/2021.acl-long.112} {Factorising
  meaning and form for intent-preserving paraphrasing}.
\newblock In \emph{Proceedings of the 59th Annual Meeting of the Association
  for Computational Linguistics and the 11th International Joint Conference on
  Natural Language Processing (Volume 1: Long Papers)}, pages 1405--1418,
  Online. Association for Computational Linguistics.

\bibitem[{Houlsby et~al.(2019)Houlsby, Giurgiu, Jastrzebski, Morrone,
  de~Laroussilhe, Gesmundo, Attariyan, and Gelly}]{houlsby2019parameter}
Neil Houlsby, Andrei Giurgiu, Stanislaw Jastrzebski, Bruna Morrone, Quentin
  de~Laroussilhe, Andrea Gesmundo, Mona Attariyan, and Sylvain Gelly. 2019.
\newblock \href {http://proceedings.mlr.press/v97/houlsby19a.html}
  {Parameter-efficient transfer learning for {NLP}}.
\newblock In \emph{Proceedings of the 36th International Conference on Machine
  Learning, {ICML} 2019, 9-15 June 2019, Long Beach, California, {USA}},
  volume~97 of \emph{Proceedings of Machine Learning Research}, pages
  2790--2799. {PMLR}.

\bibitem[{Hovy et~al.(2006)Hovy, Lin, Zhou, and Fukumoto}]{hovy2006automated}
Eduard Hovy, Chin-Yew Lin, Liang Zhou, and Junichi Fukumoto. 2006.
\newblock \href {http://www.lrec-conf.org/proceedings/lrec2006/pdf/438_pdf.pdf}
  {Automated summarization evaluation with basic elements.}
\newblock In \emph{Proceedings of the Fifth International Conference on
  Language Resources and Evaluation ({LREC}{'}06)}, Genoa, Italy. European
  Language Resources Association (ELRA).

\bibitem[{Iyyer et~al.(2018)Iyyer, Wieting, Gimpel, and
  Zettlemoyer}]{iyyer2018adversarial}
Mohit Iyyer, John Wieting, Kevin Gimpel, and Luke Zettlemoyer. 2018.
\newblock \href {https://doi.org/10.18653/v1/N18-1170} {Adversarial example
  generation with syntactically controlled paraphrase networks}.
\newblock In \emph{Proceedings of the 2018 Conference of the North {A}merican
  Chapter of the Association for Computational Linguistics: Human Language
  Technologies, Volume 1 (Long Papers)}, pages 1875--1885, New Orleans,
  Louisiana. Association for Computational Linguistics.

\bibitem[{Kazemnejad et~al.(2020)Kazemnejad, Salehi, and
  Soleymani~Baghshah}]{kazemnejad2020paraphrase}
Amirhossein Kazemnejad, Mohammadreza Salehi, and Mahdieh Soleymani~Baghshah.
  2020.
\newblock \href {https://doi.org/10.18653/v1/2020.acl-main.535} {Paraphrase
  generation by learning how to edit from samples}.
\newblock In \emph{Proceedings of the 58th Annual Meeting of the Association
  for Computational Linguistics}, pages 6010--6021, Online. Association for
  Computational Linguistics.

\bibitem[{Kumar et~al.(2020)Kumar, Ahuja, Vadapalli, and
  Talukdar}]{kumar2020syntax}
Ashutosh Kumar, Kabir Ahuja, Raghuram Vadapalli, and Partha Talukdar. 2020.
\newblock \href {https://doi.org/10.1162/tacl_a_00318} {Syntax-guided
  controlled generation of paraphrases}.
\newblock \emph{Transactions of the Association for Computational Linguistics},
  8:329--345.

\bibitem[{Lan et~al.(2017)Lan, Qiu, He, and Xu}]{lan2017continuously}
Wuwei Lan, Siyu Qiu, Hua He, and Wei Xu. 2017.
\newblock \href {https://doi.org/10.18653/v1/D17-1126} {A continuously growing
  dataset of sentential paraphrases}.
\newblock In \emph{Proceedings of the 2017 Conference on Empirical Methods in
  Natural Language Processing}, pages 1224--1234, Copenhagen, Denmark.
  Association for Computational Linguistics.

\bibitem[{Lewis et~al.(2020{\natexlab{a}})Lewis, Ghazvininejad, Ghosh,
  Aghajanyan, Wang, and Zettlemoyer}]{lewis2020pre}
Mike Lewis, Marjan Ghazvininejad, Gargi Ghosh, Armen Aghajanyan, Sida Wang, and
  Luke Zettlemoyer. 2020{\natexlab{a}}.
\newblock \href
  {https://proceedings.neurips.cc/paper/2020/hash/d6f1dd034aabde7657e6680444ceff62-Abstract.html}
  {Pre-training via paraphrasing}.
\newblock In \emph{Advances in Neural Information Processing Systems 33: Annual
  Conference on Neural Information Processing Systems 2020, NeurIPS 2020,
  December 6-12, 2020, virtual}.

\bibitem[{Lewis et~al.(2020{\natexlab{b}})Lewis, Liu, Goyal, Ghazvininejad,
  Mohamed, Levy, Stoyanov, and Zettlemoyer}]{lewis2019bart}
Mike Lewis, Yinhan Liu, Naman Goyal, Marjan Ghazvininejad, Abdelrahman Mohamed,
  Omer Levy, Veselin Stoyanov, and Luke Zettlemoyer. 2020{\natexlab{b}}.
\newblock \href {https://doi.org/10.18653/v1/2020.acl-main.703} {{BART}:
  Denoising sequence-to-sequence pre-training for natural language generation,
  translation, and comprehension}.
\newblock In \emph{Proceedings of the 58th Annual Meeting of the Association
  for Computational Linguistics}, pages 7871--7880, Online. Association for
  Computational Linguistics.

\bibitem[{Li et~al.(2018)Li, Jiang, Shang, and Li}]{li2017paraphrase}
Zichao Li, Xin Jiang, Lifeng Shang, and Hang Li. 2018.
\newblock \href {https://doi.org/10.18653/v1/D18-1421} {Paraphrase generation
  with deep reinforcement learning}.
\newblock In \emph{Proceedings of the 2018 Conference on Empirical Methods in
  Natural Language Processing}, pages 3865--3878, Brussels, Belgium.
  Association for Computational Linguistics.

\bibitem[{Li et~al.(2019)Li, Jiang, Shang, and Liu}]{li2019decomposable}
Zichao Li, Xin Jiang, Lifeng Shang, and Qun Liu. 2019.
\newblock \href {https://doi.org/10.18653/v1/P19-1332} {Decomposable neural
  paraphrase generation}.
\newblock In \emph{Proceedings of the 57th Annual Meeting of the Association
  for Computational Linguistics}, pages 3403--3414, Florence, Italy.
  Association for Computational Linguistics.

\bibitem[{Lin et~al.(2014)Lin, Maire, Belongie, Hays, Perona, Ramanan,
  Doll{\'a}r, and Zitnick}]{lin2014microsoft}
Tsung-Yi Lin, Michael Maire, Serge Belongie, James Hays, Pietro Perona, Deva
  Ramanan, Piotr Doll{\'a}r, and C~Lawrence Zitnick. 2014.
\newblock Microsoft coco: Common objects in context.
\newblock In \emph{European conference on computer vision}, pages 740--755.
  Springer.

\bibitem[{Liu et~al.(2020)Liu, Mou, Meng, Zhou, Zhou, and
  Song}]{liu2019unsupervised}
Xianggen Liu, Lili Mou, Fandong Meng, Hao Zhou, Jie Zhou, and Sen Song. 2020.
\newblock \href {https://doi.org/10.18653/v1/2020.acl-main.28} {Unsupervised
  paraphrasing by simulated annealing}.
\newblock In \emph{Proceedings of the 58th Annual Meeting of the Association
  for Computational Linguistics}, pages 302--312, Online. Association for
  Computational Linguistics.

\bibitem[{Loshchilov and Hutter(2019)}]{loshchilov2017decoupled}
Ilya Loshchilov and Frank Hutter. 2019.
\newblock \href {https://openreview.net/forum?id=Bkg6RiCqY7} {Decoupled weight
  decay regularization}.
\newblock In \emph{7th International Conference on Learning Representations,
  {ICLR} 2019, New Orleans, LA, USA, May 6-9, 2019}. OpenReview.net.

\bibitem[{Meng et~al.(2021)Meng, Ao, He, Sun, Han, Wu, Li
  et~al.}]{meng2021conrpg}
Yuxian Meng, Xiang Ao, Qing He, Xiaofei Sun, Qinghong Han, Fei Wu, Jiwei Li,
  et~al. 2021.
\newblock \href {https://arxiv.org/abs/2109.00363} {Conrpg: Paraphrase
  generation using contexts as regularizer}.
\newblock \emph{arXiv preprint arXiv:2109.00363}.

\bibitem[{Miao et~al.(2019)Miao, Zhou, Mou, Yan, and Li}]{miao2019cgmh}
Ning Miao, Hao Zhou, Lili Mou, Rui Yan, and Lei Li. 2019.
\newblock \href {https://doi.org/10.1609/aaai.v33i01.33016834} {{CGMH:}
  constrained sentence generation by metropolis-hastings sampling}.
\newblock In \emph{The Thirty-Third {AAAI} Conference on Artificial
  Intelligence, {AAAI} 2019, The Thirty-First Innovative Applications of
  Artificial Intelligence Conference, {IAAI} 2019, The Ninth {AAAI} Symposium
  on Educational Advances in Artificial Intelligence, {EAAI} 2019, Honolulu,
  Hawaii, USA, January 27 - February 1, 2019}, pages 6834--6842. {AAAI} Press.

\bibitem[{Pan and Yang(2009)}]{pan2009survey}
Sinno~Jialin Pan and Qiang Yang. 2009.
\newblock A survey on transfer learning.
\newblock \emph{IEEE Transactions on knowledge and data engineering},
  22(10):1345--1359.

\bibitem[{Papineni et~al.(2002)Papineni, Roukos, Ward, and
  Zhu}]{papineni2002bleu}
Kishore Papineni, Salim Roukos, Todd Ward, and Wei-Jing Zhu. 2002.
\newblock \href {https://doi.org/10.3115/1073083.1073135} {{B}leu: a method for
  automatic evaluation of machine translation}.
\newblock In \emph{Proceedings of the 40th Annual Meeting of the Association
  for Computational Linguistics}, pages 311--318, Philadelphia, Pennsylvania,
  USA. Association for Computational Linguistics.

\bibitem[{Park et~al.(2021)Park, Tae, Kim, Yang, Khan, Park, and
  Choo}]{park2020unsupervised}
Cheonbok Park, Yunwon Tae, TaeHee Kim, Soyoung Yang, Mohammad~Azam Khan, Lucy
  Park, and Jaegul Choo. 2021.
\newblock \href {https://doi.org/10.18653/v1/2021.acl-long.225} {Unsupervised
  neural machine translation for low-resource domains via meta-learning}.
\newblock In \emph{Proceedings of the 59th Annual Meeting of the Association
  for Computational Linguistics and the 11th International Joint Conference on
  Natural Language Processing (Volume 1: Long Papers)}, pages 2888--2901,
  Online. Association for Computational Linguistics.

\bibitem[{Pfeiffer et~al.(2021)Pfeiffer, Kamath, R{\"u}ckl{\'e}, Cho, and
  Gurevych}]{pfeiffer2020adapterfusion}
Jonas Pfeiffer, Aishwarya Kamath, Andreas R{\"u}ckl{\'e}, Kyunghyun Cho, and
  Iryna Gurevych. 2021.
\newblock \href {https://aclanthology.org/2021.eacl-main.39}
  {{A}dapter{F}usion: Non-destructive task composition for transfer learning}.
\newblock In \emph{Proceedings of the 16th Conference of the European Chapter
  of the Association for Computational Linguistics: Main Volume}, pages
  487--503, Online. Association for Computational Linguistics.

\bibitem[{Pfeiffer et~al.(2020)Pfeiffer, Vuli{\'c}, Gurevych, and
  Ruder}]{pfeiffer2020mad}
Jonas Pfeiffer, Ivan Vuli{\'c}, Iryna Gurevych, and Sebastian Ruder. 2020.
\newblock \href {https://doi.org/10.18653/v1/2020.emnlp-main.617} {{MAD-X}:
  {A}n {A}dapter-{B}ased {F}ramework for {M}ulti-{T}ask {C}ross-{L}ingual
  {T}ransfer}.
\newblock In \emph{Proceedings of the 2020 Conference on Empirical Methods in
  Natural Language Processing (EMNLP)}, pages 7654--7673, Online. Association
  for Computational Linguistics.

\bibitem[{Prakash et~al.(2016)Prakash, Hasan, Lee, Datla, Qadir, Liu, and
  Farri}]{prakash2016neural}
Aaditya Prakash, Sadid~A. Hasan, Kathy Lee, Vivek Datla, Ashequl Qadir, Joey
  Liu, and Oladimeji Farri. 2016.
\newblock \href {https://aclanthology.org/C16-1275} {Neural paraphrase
  generation with stacked residual {LSTM} networks}.
\newblock In \emph{Proceedings of {COLING} 2016, the 26th International
  Conference on Computational Linguistics: Technical Papers}, pages 2923--2934,
  Osaka, Japan. The COLING 2016 Organizing Committee.

\bibitem[{Siddique et~al.(2020)Siddique, Oymak, and
  Hristidis}]{siddique2020unsupervised}
A.~B. Siddique, Samet Oymak, and Vagelis Hristidis. 2020.
\newblock \href {https://dl.acm.org/doi/10.1145/3394486.3403231} {Unsupervised
  paraphrasing via deep reinforcement learning}.
\newblock In \emph{{KDD} '20: The 26th {ACM} {SIGKDD} Conference on Knowledge
  Discovery and Data Mining, Virtual Event, CA, USA, August 23-27, 2020}, pages
  1800--1809. {ACM}.

\bibitem[{Soh et~al.(2020)Soh, Cho, and Cho}]{soh2020meta}
Jae~Woong Soh, Sunwoo Cho, and Nam~Ik Cho. 2020.
\newblock \href {https://doi.org/10.1109/CVPR42600.2020.00357} {Meta-transfer
  learning for zero-shot super-resolution}.
\newblock In \emph{2020 {IEEE/CVF} Conference on Computer Vision and Pattern
  Recognition, {CVPR} 2020, Seattle, WA, USA, June 13-19, 2020}, pages
  3513--3522. {IEEE}.

\bibitem[{Song et~al.(2020)Song, Li, Guo, Li, Hao, Qin, and
  Zhao}]{song2020meta}
Wenfeng Song, Shuai Li, Yuting Guo, Shaoqi Li, Aimin Hao, Hong Qin, and Qinping
  Zhao. 2020.
\newblock Meta transfer learning for adaptive vehicle tracking in uav videos.
\newblock In \emph{International Conference on Multimedia Modeling}, pages
  764--777. Springer.

\bibitem[{Sun and Zhou(2012)}]{sun2012joint}
Hong Sun and Ming Zhou. 2012.
\newblock \href {https://aclanthology.org/P12-2008} {Joint learning of a dual
  {SMT} system for paraphrase generation}.
\newblock In \emph{Proceedings of the 50th Annual Meeting of the Association
  for Computational Linguistics (Volume 2: Short Papers)}, pages 38--42, Jeju
  Island, Korea. Association for Computational Linguistics.

\bibitem[{Wang et~al.(2022)Wang, Xu, Sun, Hu, Tao, Geng, and
  Jiang}]{wang2022promda}
Yufei Wang, Can Xu, Qingfeng Sun, Huang Hu, Chongyang Tao, Xiubo Geng, and
  Daxin Jiang. 2022.
\newblock \href {https://arxiv.org/abs/2202.12499} {Promda: Prompt-based data
  augmentation for low-resource nlu tasks}.
\newblock \emph{arXiv preprint arXiv:2202.12499}.

\bibitem[{Wang et~al.(2019)Wang, Dai, P{\'{o}}czos, and
  Carbonell}]{wang2019characterizing}
Zirui Wang, Zihang Dai, Barnab{\'{a}}s P{\'{o}}czos, and Jaime~G. Carbonell.
  2019.
\newblock \href {https://doi.org/10.1109/CVPR.2019.01155} {Characterizing and
  avoiding negative transfer}.
\newblock In \emph{{IEEE} Conference on Computer Vision and Pattern
  Recognition, {CVPR} 2019, Long Beach, CA, USA, June 16-20, 2019}, pages
  11293--11302. Computer Vision Foundation / {IEEE}.

\bibitem[{West et~al.(2021)West, Lu, Holtzman, Bhagavatula, Hwang, and
  Choi}]{west2020reflective}
Peter West, Ximing Lu, Ari Holtzman, Chandra Bhagavatula, Jena~D. Hwang, and
  Yejin Choi. 2021.
\newblock \href {https://doi.org/10.18653/v1/2021.acl-long.114} {Reflective
  decoding: Beyond unidirectional generation with off-the-shelf language
  models}.
\newblock In \emph{Proceedings of the 59th Annual Meeting of the Association
  for Computational Linguistics and the 11th International Joint Conference on
  Natural Language Processing (Volume 1: Long Papers)}, pages 1435--1450,
  Online. Association for Computational Linguistics.

\bibitem[{Yin et~al.(2022)Yin, Li, and Li}]{yin2022learning}
Haiyan Yin, Dingcheng Li, and Ping Li. 2022.
\newblock Learning to selectively learn for weakly supervised paraphrase
  generation with model-based reinforcement learning.
\newblock In \emph{Proceedings of the 2022 Conference of the North American
  Chapter of the Association for Computational Linguistics: Human Language
  Technologies}, pages 1385--1395.

\bibitem[{Zhao et~al.(2008)Zhao, Niu, Zhou, Liu, and Li}]{zhao2008combining}
Shiqi Zhao, Cheng Niu, Ming Zhou, Ting Liu, and Sheng Li. 2008.
\newblock \href {https://aclanthology.org/P08-1116} {Combining multiple
  resources to improve {SMT}-based paraphrasing model}.
\newblock In \emph{Proceedings of ACL-08: HLT}, pages 1021--1029, Columbus,
  Ohio. Association for Computational Linguistics.

\end{thebibliography}
\bibliographystyle{acl_natbib}

\clearpage
\appendix
\section{Appendix}
\subsection{Datasets Details}
\textbf{Quora} Quora includes 260K negative and 140 positive Quora question paraphrase pairs.
We only use positive pairs and follow the same setting in ~\citet{li2017paraphrase, kazemnejad2020paraphrase,ding2021learning} and randomly sample 100K, 30K, 3K parallel sentences for training, test, and validation, respectively. 
Low-resource settings use the same validation and test set, but the training set size is reduced.

\noindent \textbf{MSCOCO} MSCOCO~\citep{lin2014microsoft} contains about 500K human annotated captions of over 120K images, i.e. each image contains five captions from five different annotators. 
We follow the standard data split according to ~\citet{lin2014microsoft, liu2019unsupervised,ding2021learning}. 

\noindent \textbf{WikiAnswer} WikiAnswer~\citep{fader2013paraphrase} contains approximately 18 million paraphrases that are word-aligned question pairs. 
We only use this dataset as the source task of meta-training, and follow the standard data split according to ~\citet{li2019decomposable, liu2019unsupervised,siddique2020unsupervised}.

\noindent \textbf{Twitter} The twitter URL paraphrasing corpus is built by ~\citet{lan2017continuously} for paraphrase identification. We follow the setting in ~\citet{li2017paraphrase}, ~\citet{kazemnejad2020paraphrase} and ~\citet{siddique2020unsupervised}. 

The detailed dataset statistics are summarized in Table~\ref{dataset-statistics-table} .
\begin{table}[htbp]
\centering
\scalebox{0.7}{
\begin{tabular}{lrrrrrrr}
\hline
\multirow{2}{*}{\textbf{Datasets}} & 
\multicolumn{3}{c}{\textbf{Train} } & 
\multirow{2}{*}{\textbf{Valid}}    & 
\multirow{2}{*}{\textbf{Test}}     & 
\multirow{2}{*}{\textbf{Vocab}} \\  
 \cline{2-4}
& \textbf{S.} & \textbf{L.R.} & \textbf{U.S.} &     &   &  & \\
\hline
Quora    & 100K    & 3K    & 0K      & 3K       & 30K      & 8K     \\
MSCOCO   & 110K    & 10K    & 0K      & 10K      & 40K     & 8K     \\
Twitter  & 110K    & 1K    & 0K      & 1K       & 5K       & 10K    \\
WikiAnswer & 500k  & 20k   & 0k      & 20k      & 6k      &  8k     \\
\hline
\end{tabular}}
\caption{
  \label{dataset-statistics-table} 
  Statistics of datasets. S./L.R./U.S. denote supervised, low-resource, and unsupervised settings, respectively.}
  \vskip -1em
\end{table}
\subsection{Evaluation Details}
To make a fair and comprehensive assessment, we follow the same experiment setting of each comparison work~\citep{li2017paraphrase,liu2019unsupervised,ding2021learning} and conduct the comparison respectively. For data preprocessing, all the sentences are lower cased, and truncate all sentences to up to 20 words. <s> and </s> are spliced to the front and back end of the sentence as start and end markers. 

For evaluation metrics, we use BLEU, i-BLEU and ROUGE that have been widely used in the previous work to measure the quality of the paraphrases. The i-BLUE aims to measure the diversity of expression in the generated paraphrases by penalizing copying words from input sentences. Specifically, we follow the unsupervised paraphrase generation baselines and set the balancing parameter $\alpha$ = 0.9. 

\subsection{Implementation}
\label{sec:implementation}
Our experiments were conducted with PyToch on NVIDIA Tesla V100 16GB GPU. 
Following the comparison methods, we used BART-large as the pre-trained language model and use its pre-trained parameters.  
For adapter modules, the hidden size is 128. 
For meta-training, unless otherwise specified, a meta batch includes 3 tasks, and the batch size of each task is 10. 
Both basic learners and meta learners use the AdamW~\citep{loshchilov2017decoupled} optimizer for optimization, and the learning rate is set by grid search in 1e-5, 5e-5, 1e-6 and 5e-6.
The internal gradient step size is 4, and the whole model has enough step size for training. 
For meta verification, we use a corpus excluded from the source task and the target task. 
For fine-tuning, we use validation set to select the best model for metrics calculation. 

\end{document}